\documentclass[11pt]{article}
\usepackage[margin=1in]{geometry}
\usepackage[T1]{fontenc}
\usepackage[utf8]{inputenc}
\usepackage{lmodern}
\usepackage{microtype}
\usepackage{amsmath,amssymb}
\usepackage{graphicx}
\usepackage{booktabs}
\usepackage{caption}
\usepackage[section]{placeins}
\usepackage[natbibapa]{apacite}
\usepackage{authblk}
\usepackage{orcidlink}

\usepackage{lineno}
\usepackage{xcolor}
\usepackage{hyperref}
\hypersetup{
    colorlinks=true,
    linkcolor=blue!60!black,
    citecolor=blue!60!black,
    urlcolor=blue!60!black,
}

\title{\textbf{How LLMs See Creativity: Zero-Shot Scoring of Visual Creativity with Interpretable Reasoning}}
\author[1]{William Orwig\,\orcidlink{0000-0002-9913-6391}}
\author[2,*]{Roger E. Beaty\,\orcidlink{0000-0001-6114-5973}}
\affil[1]{Harvard University, Department of Psychology, Cambridge, MA, USA}
\affil[2]{Pennsylvania State University, Department of Psychology, University Park, PA, USA}
\affil[*]{Corresponding Author: Roger E. Beaty (email: rebeaty@psu.edu)}
\date{}

\begin{document}
\maketitle

\begin{abstract}
Evaluating the originality of visual images poses enduring challenges for creativity assessment. Automated scoring using AI models has proven effective in the verbal domain, yet key questions remain about evaluating visual creativity and understanding how models arrive at their ratings. The present research asks whether multimodal large language models (LLMs) can serve as judges of visual creativity zero-shot---without any fine-tuning or examples of human ratings---and whether their ``reasoning'' output offers an interpretable window into their evaluation process. We tested six multimodal LLMs (Gemini 3 Flash, Gemma 4 31B IT, GPT-5.4 Mini, GLM-5v Turbo, Kimi K2.5, and Qwen 3.6 Plus) on 992 AI-generated images (based on human-written prompts) and 1,500 hand-drawn sketches scored for creativity by human raters. In Study 1, all models showed substantial alignment with human creativity ratings on both datasets (r = .57--.68 on AI-generated images; r = .29--.68 on sketches). In Study 2, we analyzed the step-by-step reasoning processes of three LLMs evaluating the same images and drawings. Although reasoning made model evaluations interpretable---showing what they attend to, how they balance originality vs. quality, and how they justify their ratings---reasoning did not improve alignment with human ratings. In sum, our findings indicate that multimodal LLMs can match human judgments of visual creativity without any additional training, and that their reasoning reveals how AI models evaluate creativity. An open scoring app implementing this pipeline is available at \url{https://review-visual-eval-scoring.hf.space}.

\vspace{0.5em}
\end{abstract}

\section*{Introduction}
As LLMs now produce large volumes of open-ended outputs, the bottleneck in research is often no longer generation, but evaluation. This problem is especially acute for creative products, where quality is subjective, multidimensional, and imperfectly captured by existing benchmarks. The rise of generative AI has shifted attention toward forms of co-creativity between humans and AI in which systems can support idea generation, selection, and production across many domains \citep{rafner2023creativity}. In this transitional moment, several empirical questions emerge: Can LLMs support human evaluations? How much instruction and training data are needed to achieve a reasonable correspondence between human and LLM ratings? And critically, what are the latent, unobserved processes that support LLM assessments of creativity?

Creative performance has traditionally been assessed through human judgment, with the Consensual Assessment Technique remaining a standard approach for evaluating ideas and products in ways that are sensitive to domain knowledge and expertise \citep{amabile1982cat}. Related subjective scoring methods have also been shown to yield reliable and valid judgments in divergent thinking research when multiple raters evaluate responses independently \citep{silvia2008subjective}. At the same time, these methods are labor-intensive and difficult to scale, especially when studies involve thousands of responses. This tension between psychological validity and scale has motivated growing interest in automated approaches to creativity assessment; platforms that combine online administration with automated scoring of creativity tasks \citep{patterson2025cap} have made large-scale assessment more accessible and set the stage for model-based evaluative approaches.

Recent work in computational psychometrics has sought to develop automated methods for creativity scoring on the alternate uses task (AUT), the benchmark measure of divergent thinking. Early methods provided a proxy for originality by quantifying how remote a response is from a prompt or from other concepts in semantic space (i.e., SemDis; \citealp{beaty2021semdis}). More recent LLM-based approaches have trained models on large datasets of human-scored examples \citep{organisciak2023beyond}, yielding strong alignment (r = .81) with human judgments. This work has been extended to other domains of creativity, including metaphor production \citep{distefano2025metaphor}, narrative writing \citep{orwig2024language}, and figural creativity \citep{patterson2024audra}. In the figural domain specifically, \citet{cropley2022cnn} trained a convolutional neural network on the Test of Creative Thinking - Drawing Production and achieved .83--.94 classification accuracy; \citet{cropley2025tctdp} extended this with a large-scale open-source pipeline. Similarly, \citet{acar2025figural} applied computer-vision classifiers to figural Torrance tests of divergent thinking (r = .54--.85). Taken together, these studies suggest that model-based scoring can recover meaningful variance in human creativity judgments across several tasks.

Even without task-specific training, general-purpose LLMs can score verbal creativity surprisingly well. Zero-shot refers to model-based evaluations in which only a prompt and response are provided; similarly, few-shot provides the model with a limited set of example responses/human ratings in the prompt (i.e., in-context learning; \citealp{dong2024icl}). \citet{organisciak2023beyond} found that prompt-based GPT-4 scoring of AUT responses achieved correlations of r = .53 in zero-shot and r = .66 in few-shot conditions. \citet{saretzki2025german} extended zero-shot GPT-4 scoring to German-language AUT responses and reported an average correlation of r = .55 with human ratings, with considerable variability across items and studies (r = .34--.70). Relatedly, \citet{luchini2025cps} applied few-shot prompting to a creative problem-solving task and found correlations with human originality ratings ranging from r = .11 to .66. Although zero-shot scoring has shown promise for some verbal creativity tasks, it is unknown whether zero-shot evaluation extends to the visual domain.

Most current LLMs are multimodal, accommodating both images and text, and their ability to serve as zero-shot judges has been documented for text-based tasks \citep{chiang2023llm_eval,zheng2023judge}. A recent development is reasoning-capable LLMs that produce explicit chains of reasoning before committing to an answer \citep{wei2022cot,guo2025deepseek}. In extended-reasoning mode, the chain is returned separately from the model's final answer, revealing how it ``thinks'' about the user's request before outputting a response. For creativity assessment, this offers an opportunity to glean new insights into the underlying processes supporting creative evaluation. 

The present research evaluates general-purpose multimodal LLMs as zero-shot judges of visual creativity at two levels of analysis: behavioral alignment with human ratings and the evaluative content of the models' reasoning chains. We tested six multimodal LLMs from five different developers on two visual creativity datasets---992 AI-generated images from human-written prompts  \citep{orwig2026aiart} and 1,500 hand-drawn sketches \citep{patterson2024audra}---with no task-specific training, no exemplars, and no fine-tuning. Study 1 tests all six models against human ratings on both datasets, assessing their alignment beyond simple visual complexity. Study 2 collects the reasoning chains produced by three reasoning-capable LLMs, classifying each reasoning sequence into discrete steps (e.g., image perception, originality evaluation, rating justification) and by testing whether the models can correctly identify what each drawing actually depicts---a particular challenge with coarse, hand-drawn sketches. Together, these studies aimed to expand multimodal automated creativity scoring and provide insight into how AI models evaluate creativity. To make the scoring procedure usable on new images, we also share an open web app that runs the same pipeline on any user-supplied stimulus (Appendix~A).

\section*{Study 1: Zero-Shot Visual Creativity Ratings}
\subsection*{Method}
\subsubsection*{Datasets}
The present research used two visual creativity datasets that differ in stimulus type. Dataset 1 (``AI-generated images'' hereafter) consisted of 992 images generated via DALL-E 3 from brief participant-written creative word sets \citep{orwig2026aiart}. Three trained raters independently evaluated each image for creativity on a 1-to-5 scale (1 = very uncreative; 5 = very creative), blind to the generating prompts. Inter-rater reliability was adequate (ICC = .66). The criterion score for each image was the continuous average of the three ratings (M = 2.39, SD = 0.80; values fall on 1/3-point increments across the 1--5 range). The full sample of 992 images was used in all analyses.

Dataset 2 (``human-drawn sketches'' hereafter) consisted of drawings collected via the AuDrA platform \citep{patterson2024audra}, in which participants were given an incomplete starting shape and asked to extend it into a creative drawing. The full dataset includes 11,075 drawings from 14 distinct starting shapes, each rated by approximately 50 trained raters using a planned missing design. Human creativity scores were derived using a Judge Response Theory (JRT) graded response model that accounts for rater severity and measurement error \citep{myszkowski2019jrt}. For comparability with the 1-to-5 ratings of \citet{orwig2026aiart}, the rescaled integer 1-to-5 form is also reported where indicated.

\begin{figure}[!ht]
\centering
\includegraphics[width=\linewidth]{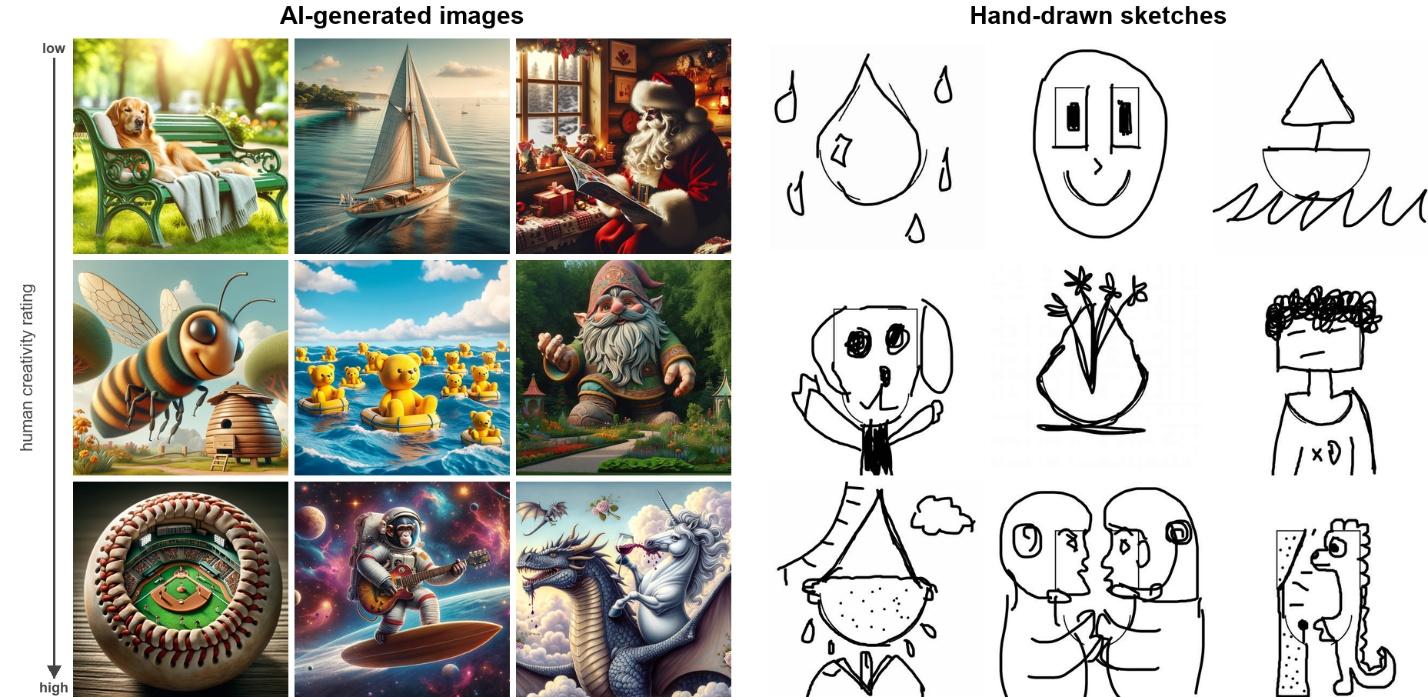}
\caption{ Example stimuli, sorted top-to-bottom from lowest to highest mean human creativity rating. (A) AI-generated images produced with DALL-E 3 from participant-written word sets (\citealp{orwig2026aiart}; N = 992 in full sample). (B) Hand-drawn sketches extending an incomplete starting shape (\citealp{patterson2024audra}; N = 1,500 in present subsample).}\label{fig:examples}
\end{figure}

To balance within-item consistency against between-item variability in Dataset 2, we sampled drawings from three distinct abstract starting shapes (shapes 4, 11, and 12 in the full dataset). Within each shape, we randomly sampled 500 drawings (random seed 42), yielding a final sample of 1,500 drawings (rescaled integer ratings: M = 2.95, SD = 0.71; range 1--5). This design holds the starting stimulus constant within each shape so that variation in human and model ratings reflects how different participants solved the same prompt while also capturing item-level variability. Shape membership was retained as a metadata variable for subgroup analyses. Figure~\ref{fig:examples} shows representative examples from each dataset across the range of human creativity ratings.

\subsubsection*{Models}
We tested six multimodal LLMs from five developers, capable of processing image inputs (Table~\ref{tab:models}). All six models operated in a comparable, single-pass generative mode for Study 1; reasoning chains from the three reasoning-capable models were collected separately in Study 2.

\begin{table}[!htbp]
\centering
\small
\caption{Multimodal LLMs evaluated in the present research.}\label{tab:models}
\begin{tabular}{@{}lll@{}}
\toprule
Model & Developer & Accessed \\
\midrule
Gemini 3 Flash & Google & March 2026 \\
Gemma 4 31B IT & Google & April 2026 \\
GPT-5.4 Mini & OpenAI & April 2026 \\
GLM-5v Turbo & Z-AI & April 2026 \\
Kimi K2.5 & Moonshot AI & April 2026 \\
Qwen 3.6 Plus & Alibaba & April 2026 \\
\bottomrule
\end{tabular}
\end{table}

Note. Exact API identifiers (in order): \texttt{gemini-3-flash-preview}, \texttt{gemma-4-31b-it}, \texttt{gpt-5.4-mini-20260317}, \texttt{glm-5v-turbo}, \texttt{kimi-k2.5}, \texttt{qwen3.6-plus}. Gemini 3 Flash was accessed directly through the Google AI API (\texttt{genai} SDK); all other models were accessed through OpenRouter. GLM-5v Turbo, Kimi K2.5, and Qwen 3.6 Plus support extended-reasoning mode, which was disabled in Study 1 and enabled in Study 2.

\subsubsection*{Procedure}
Each model received a single image and a text prompt instructing it to rate the image's creativity on a 1-to-5 scale. All requests used temperature = 0 to ensure deterministic outputs. Responses were parsed for the first integer between 1 and 5, and failed parses were retried up to four times with exponential backoff. API requests were issued asynchronously with concurrency limits set to respect provider rate limits.

For the AI-generated images, the prompt was deliberately sparse to mirror the brevity of the original human-rater instructions in \citet{orwig2026aiart}:

\begin{quote}
You are evaluating the creativity of AI-generated images. Rate the creativity on a scale of 1 to 5, where 1 is very uncreative and 5 is very creative. Use the ENTIRE 1 to 5 scale. Provide only a single number as your rating.
\end{quote}

For the hand-drawn sketches, the prompt was adapted from the human-rater instructions in \citet{patterson2024audra}. The LLM prompt similarly emphasized originality over artistic quality and provided context about the incomplete-shape drawing procedure:

\begin{quote}
You are evaluating the creativity of drawings created by various people in research studies (not necessarily artists). Rate the creativity on a scale of 1--5, where 1 is not at all creative and 5 is very creative. Use the ENTIRE 1 to 5 scale for the rating. Don't hesitate to use extreme values when appropriate. Focus on the originality of the idea, not the artistic quality. The drawing was created using a starting image of an incomplete shape, which was incorporated into the drawings. Provide only a single number between 1 and 5 as your rating, where: 1 = Not at all creative; 2 = Slightly creative; 3 = Moderately creative; 4 = Very creative; 5 = Extremely creative.
\end{quote}

\subsubsection*{Edge Density}
To account for visual-complexity confounds, where both humans and AI models give higher ratings to more elaborate responses \citep{domanti2026elaboration}, we extracted edge density from each image, defined as the proportion of pixels exceeding a Sobel-filter gradient threshold of 0.1 (range 0--1). Edge density indexes the amount of visual content in an image and is known to predict human creativity ratings on figural drawing tasks, where elaboration and originality are correlated \citep{patterson2024audra}. Here, we use edge density as a covariate for partial correlations, not as a dependent variable.

\subsubsection*{Analytic Approach}
For each model, we computed the Pearson correlation between its ratings and the human criterion scores, alongside the partial correlation controlling for edge density. Rating distributions were summarized as means, standard deviations, and proportions of ratings at each scale point to characterize systematic biases relative to human raters. For the hand-drawn-sketch sample, we additionally report per-shape correlations to characterize item-level variability.

\subsection*{Results}
\begin{figure}[!ht]
\centering
\includegraphics[width=\linewidth]{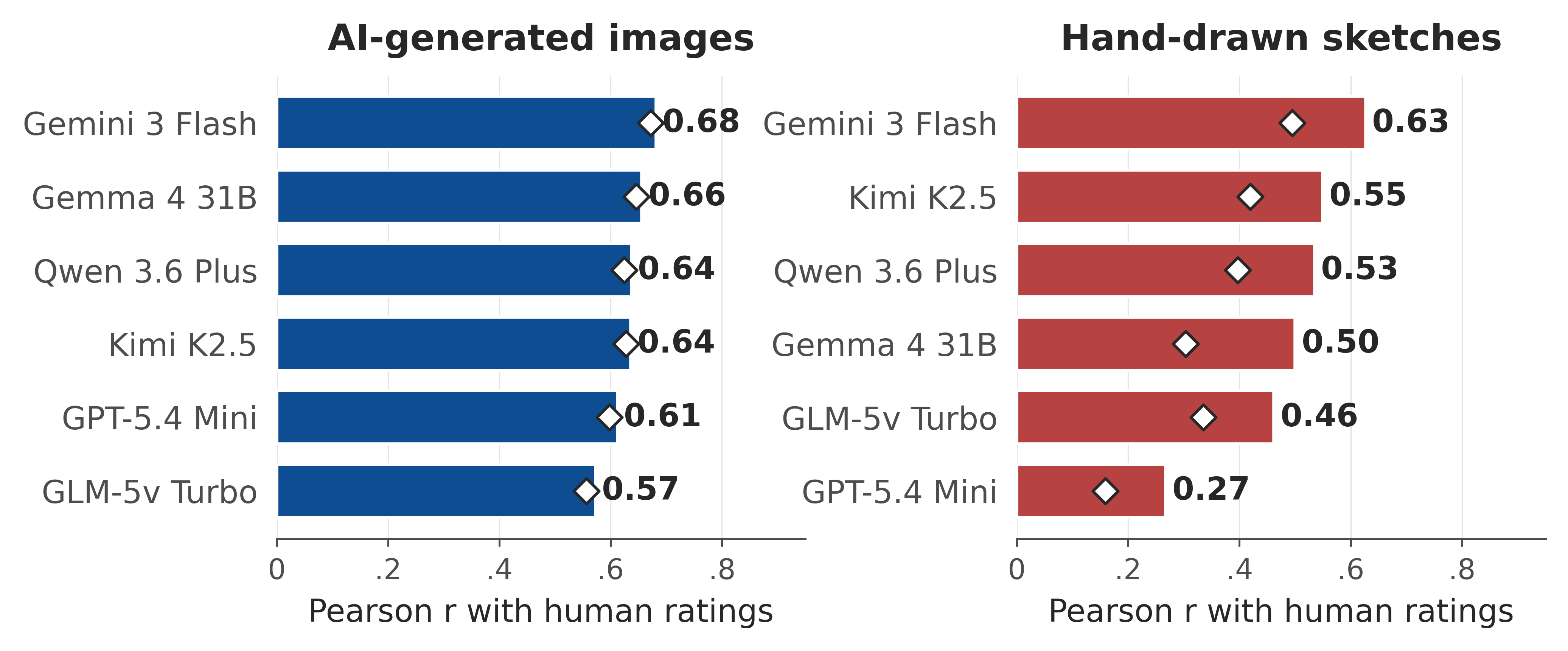}
\caption{Model–human alignment by dataset. For each of the six LLMs, bar length shows the bivariate Pearson r with human ratings and the white diamond shows the partial r controlling for edge density. Models are ranked by Pearson r within each panel. All models align well with humans on AI-generated images. On hand-drawn sketches, controlling for edge density lowers alignment, showing that part of their agreement with humans reflects visual complexity rather than originality.}\label{fig:study1-alignment}
\end{figure}

\begin{figure}[!ht]
\centering
\includegraphics[width=\linewidth]{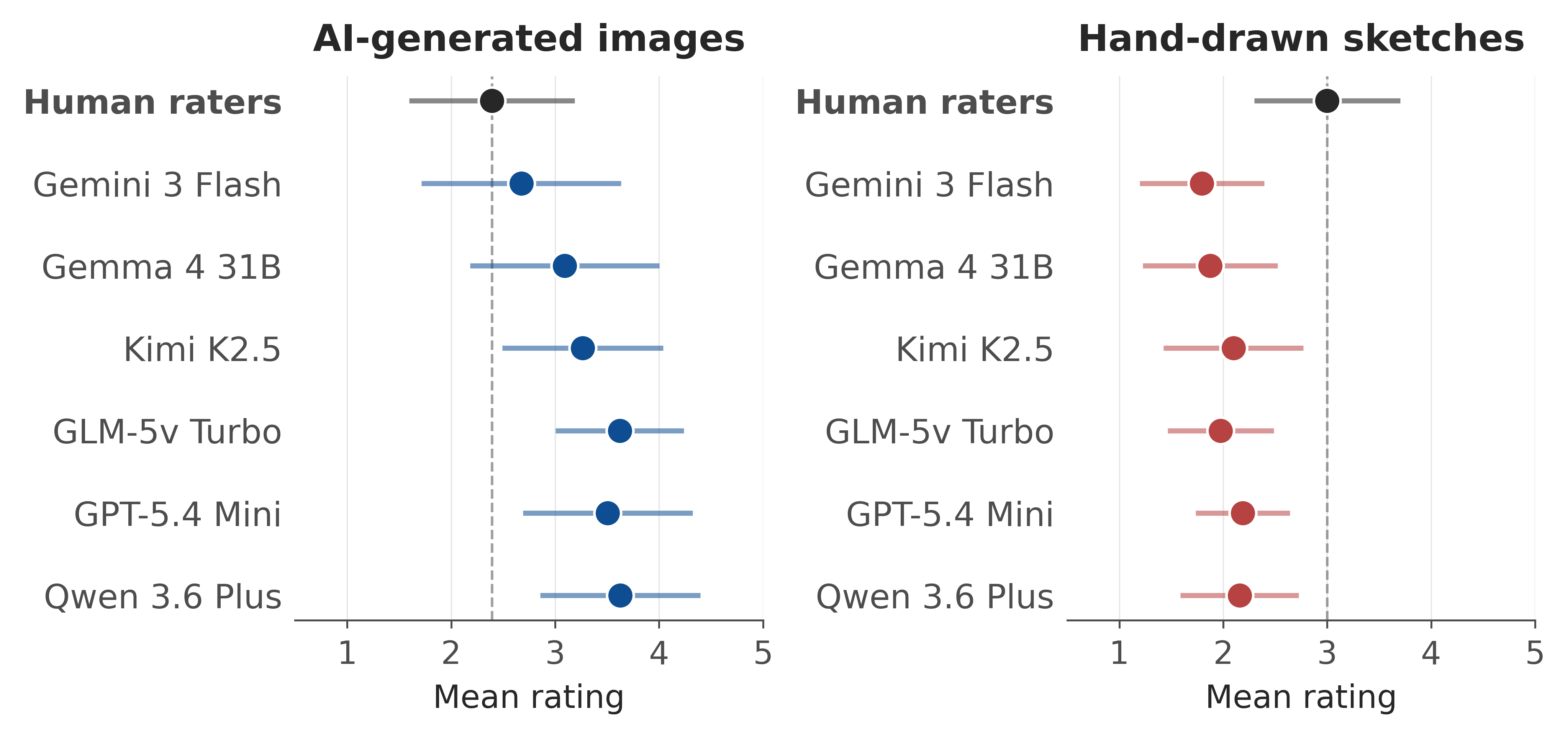}
\caption{Mean creativity rating per source on each dataset. Each dot is one source's mean (humans in black, LLMs in color); whiskers span ±1 SD and the dashed line marks the human mean. Rows are ordered identically across panels. The bias reverses by dataset: every model rates AI-generated images more leniently than humans and hand-drawn sketches more harshly.}\label{fig:rating-distributions}
\end{figure}

\paragraph{Multimodal LLMs align with human creativity ratings of AI-generated images.} We first tested the extent to which zero-shot LLM ratings correlate with human creativity ratings on the AI-generated images. Results showed moderate-to-strong correlations (Figure~\ref{fig:study1-alignment}). Gemini 3 Flash showed the strongest alignment (r = .68), followed by Gemma 4 31B IT (r = .66), Kimi K2.5 (r = .64), Qwen 3.6 Plus (r = .64), GPT-5.4 Mini (r = .61), and GLM-5v Turbo (r = .57). Edge density was only weakly correlated with human ratings on this dataset (r = .17), and partial correlations controlling for edge density were nearly identical to the bivariate correlations (partial r = .56--.67), suggesting that LLM--human alignment for AI-generated images is not attributable to shared sensitivity to low-level visual complexity. Notably, models showed a systematic leniency bias relative to human raters. Human ratings concentrated at the lower end of the scale (M = 2.39, SD = 0.80; 42\% of images rated at or below 2.0), whereas model ratings were shifted upward (M = 2.67--3.63 across models). Most models rarely or never assigned a rating of 1 (Figure~\ref{fig:rating-distributions}), compressing their judgments toward the upper-middle of the scale.

\paragraph{LLM-human alignment extends to hand-drawn sketches.} We next examined whether this alignment generalized to a fundamentally different stimulus type. Overall, most models retained strong alignment with human ratings, with notable variability (Figure~\ref{fig:study1-alignment}). Gemini 3 Flash again showed the strongest alignment (r = .68), followed by Kimi K2.5 (r = .60), Qwen 3.6 Plus (r = .57), Gemma 4 31B (r = .55), GLM-5v Turbo (r = .49), and GPT-5.4 Mini (r = .29). Interestingly, the relative ranking of models shifted between datasets: Kimi K2.5 and Qwen 3.6 Plus moved up from middle ranks on AI-generated images to second and third on drawings, while Gemma 4 31B dropped from second to fourth.

\paragraph{Alignment persists beyond visual complexity.} Regarding edge density, we found a large overlap with human ratings (r = .64) on this dataset, substantially stronger than AI-generated images. This elaboration effect is expected, however, given prior analyses of these line drawings \citep{patterson2024audra}. After partialing out edge density, LLM--human correlations were attenuated, but the rank order was unchanged, and Gemini 3 Flash retained the strongest alignment (partial r = .56; range across models = .17--.56). Notably, after accounting for edge density, GPT-5.4 Mini, a frontier closed-source model, showed poor alignment to human ratings, while partial correlations for open-source models (Kimi and Qwen) remained large (Figure~\ref{fig:study1-alignment}). Nevertheless, the substantial residual correlation for some LLMs highlights their ability to detect features judged as creative by humans, beyond simply their complexity.

\paragraph{LLMs prefer AI-generated images.} In contrast to the leniency bias observed on AI-generated images, models showed a negative bias on hand-drawn sketches. We found that human ratings centered on the scale midpoint of 3 (M = 2.95, SD = 0.71), whereas model ratings were shifted downward (M = 1.79--2.19 across models), with most models assigning the majority of ratings to 1 or 2 (Figure~\ref{fig:rating-distributions}). Thus, models inflated ratings on polished AI-generated images and deflated ratings on sparse, hand-drawn sketches, suggesting that the models penalize visual simplicity despite prompt instructions to focus on the originality of the idea expressed.

\paragraph{Items show variable alignment for human-drawn sketches.} Because the hand-drawn-sketch sample included 500 drawings each from three distinct starting shapes (4, 11, 12), we could compute the model--human correlations separately for each shape (Table~\ref{tab:shape-correlations}). We found shape 4 elicited consistently stronger alignment than shapes 11 and 12 across nearly every LLM. The effect was largest for GLM-5v Turbo, whose Pearson r dropped from .66 on shape 4 to .36 on shape 11, and for Kimi K2.5 and Qwen 3.6 Plus, both of which dropped roughly .10--.15 from shape 4 to shape 11. Gemini 3 Flash and Gemma 4 31B were the most stable across items, with cross-shape ranges of about .03--.07. GPT-5.4 Mini was uniformly weak across shapes (r = .26--.34). However, we found edge density predicted human ratings consistently across shapes, so the item effects likely reflect model performance rather than differences in stimulus rateability.

\begin{table}[!htbp]
\centering
\small
\caption{Model--human correlations by starting shape in human-drawn sketches (N = 500 per shape, against continuous JRT theta).}\label{tab:shape-correlations}
\begin{tabular}{@{}llccc@{}}
\toprule
Model & Shape 4 & Shape 11 & Shape 12 & Pooled \\
\midrule
Gemini 3 Flash & .70 & .67 & .67 & .68 \\
Kimi K2.5 & .69 & .58 & .55 & .60 \\
Qwen 3.6 Plus & .65 & .50 & .56 & .57 \\
Gemma 4 31B & .59 & .57 & .51 & .55 \\
GLM-5v Turbo & .66 & .36 & .42 & .49 \\
GPT-5.4 Mini & .30 & .26 & .34 & .29 \\
\bottomrule
\end{tabular}
\end{table}

\section*{Study 2: Reasoning Chains in Visual Creativity Judgments}
Study 1 established that multimodal LLMs can predict human creativity ratings across two very different datasets, with moderate to strong alignment despite no additional training (i.e., zero-shot). Gemini 3 Flash showed the strongest alignment on both datasets, with open-source models (Kimi K2.5, Qwen 3.6 Plus) outperforming a leading proprietary LLM (GPT-5.4 Mini). In Study 2, we explore how models arrive at their ratings. To this end, we analyze the explicit reasoning chains produced by the three reasoning-capable models from Study 1. These chains are made visible separately from the final rating, giving direct access to what the model notices in the images and how it ``thinks'' about creativity. We used these reasoning chains to address four questions: (1) Does reasoning improve alignment with human ratings (compared to a baseline of no reasoning)? (2) Does the reasoning chain itself carry meaningful information beyond the rating (can we predict the final rating from the reasoning content alone)? (3) What do the models actually consider when judging creativity? And (4) Does correctly identifying what a drawing depicts matter for matching human ratings?

\subsection*{Method}
\paragraph{Reasoning chain collection and validation.} We re-ran the three reasoning-capable models (GLM-5v Turbo, Kimi K2.5, and Qwen 3.6 Plus) on the full Study 1 samples with reasoning enabled, recording each model's final 1-to-5 rating and the full reasoning chain it produced. Rating prompts were identical to Study 1.

A central concern with reasoning chains is that they may not reflect the model's actual evaluation, i.e., the chain could simply be post-hoc rationalization, with the rating decided independently \citep{chiang2023llm_eval,zheng2023judge}. We thus asked a separate LLM (GPT-5.4 Nano) to guess each rating from the chain alone, with no access to the image. To prevent this prediction model from simply pattern-matching on explicit candidate-rating phrases (e.g., "I'd give this a 3"), we removed all numeric digits from the chain text before passing it to the predictor. The predicted rating was scored against (a) the original LLM rating that produced the chain, and (b) the human rating. If the chains contain the evaluative signal that drives the rating, a text-only model should be able to reconstruct it; if the chains are simply post-hoc rationalizations, the predicted ratings should be uncorrelated with the original ones.

\paragraph{Coding the reasoning process.} To characterize what models attend to when judging creativity, we decomposed each reasoning chain into its constituent sentences and classified each sentence into one of four evaluative dimensions (or a residual Other category). The four categories follow the perception -> reasoning -> integration decomposition from vision-language chain-of-thought analysis \citep{avogaro2026sparc,jiang2025vlmr3}. This taxonomy was operationalized for the creativity-judgment task as: Perception (describes visible content, enumerates what the model sees, or restates the task; no evaluation), Originality (judges novelty, surprise, conventionality, cliche, rarity, or references how common similar content is), Quality (judges visual quality, composition, rendering, technique, detail, or polish), and Justification (reasons explicitly about the 1-to-5 scale, anchors the rating, or commits to a final rating with brief justification). The fifth category, Other, captured transitions, meta-commentary, and any sentence that did not fit the four evaluative categories.

\begin{figure}[!ht]
\centering
\includegraphics[width=\linewidth]{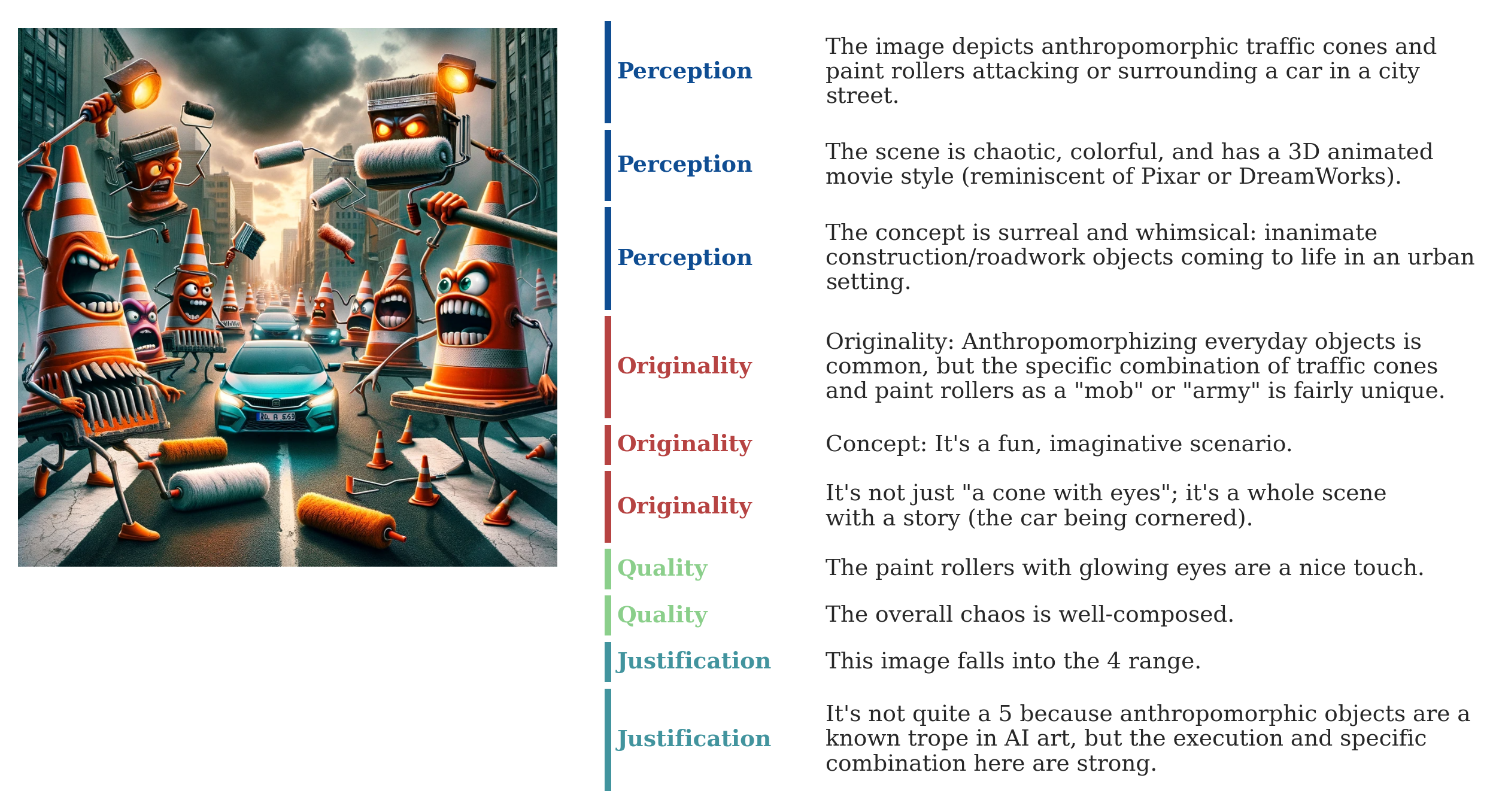}
\caption{Example reasoning chain for an AI-generated image (\citealp{orwig2026aiart}; image 77), from GLM-5v Turbo with reasoning enabled. Sentences are color-coded by evaluative category: Perception, Originality, Quality, Justification or Other. Sentences are verbatim; brief task-restatement and transition lines are omitted.}\label{fig:chain-ai}
\end{figure}

\begin{figure}[!ht]
\centering
\includegraphics[width=\linewidth]{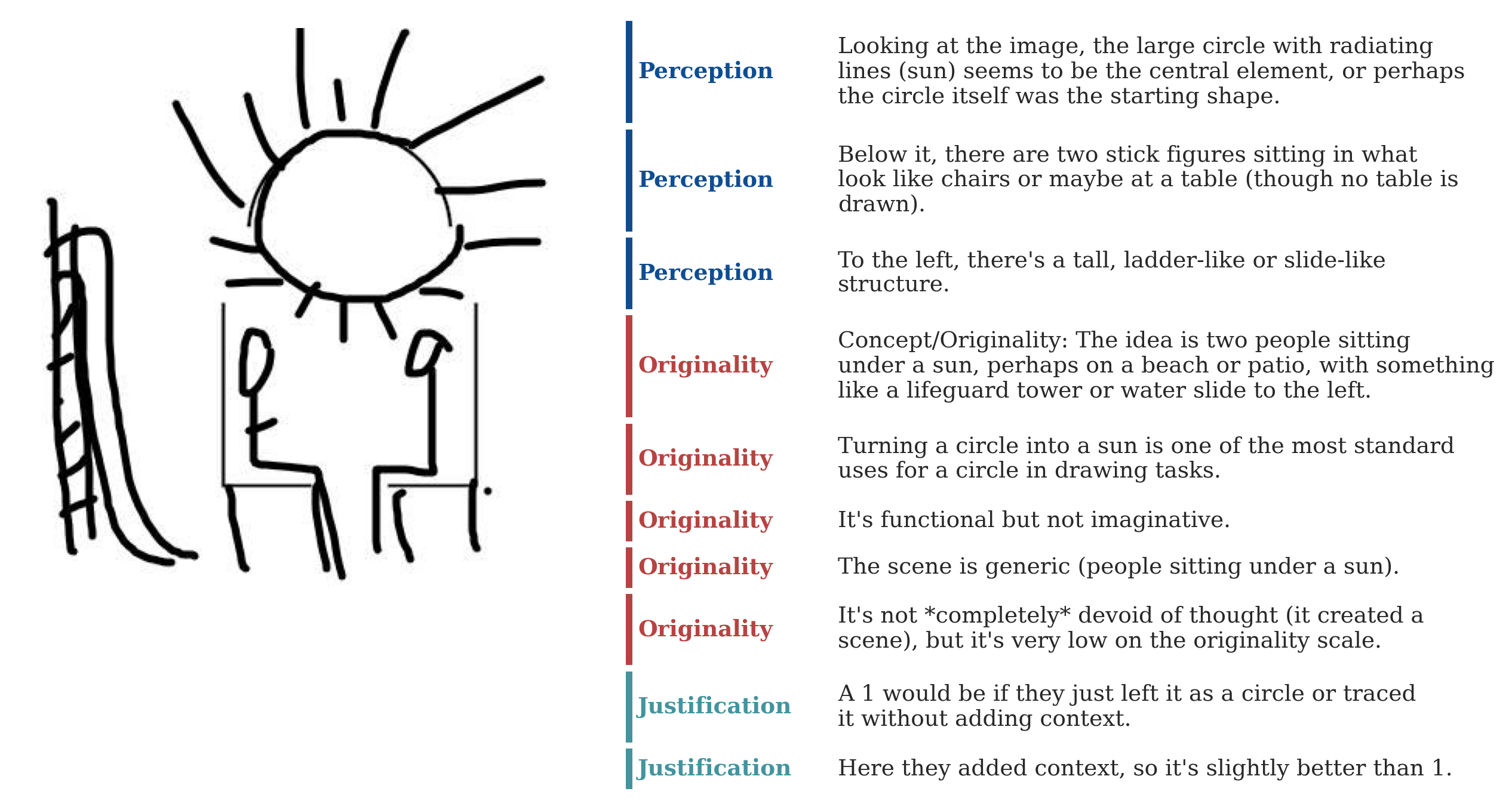}
\caption{Example reasoning chain for a hand-drawn sketch from GLM-5v Turbo with reasoning enabled. Color-coding as in Figure~\ref{fig:chain-ai}. Unlike the AI-image chain, Quality sentences are absent and Perception sentences dominate---a shift that recurs across all three reasoning-capable LLMs (Figure~\ref{fig:process-distribution}). Sentences are reproduced verbatim; a brief task-restatement line has been omitted for presentation.}\label{fig:chain-sketch}
\end{figure}

Sentences were extracted by splitting each chain on period, question-mark, and exclamation-point terminators and on double newlines, then dropping fragments shorter than 25 characters and number-only fragments. The three judges were Mistral Small (mistralai/mistral-small-2603), Grok 4.1 Fast (x-ai/grok-4.1-fast), and MiMo v2 Pro (xiaomi/mimo-v2-pro, with reasoning explicitly disabled), all accessed via OpenRouter at temperature = 0. We selected this set of judges to use models from developers not represented in our Study 1 model set, providing independence between the rated models and the coding pipeline. To make the coding pass tractable across 562,053 sentences, each API call presented 15 numbered sentences and elicited a JSON response mapping each number to one category. A pilot comparison against one-call-per-sentence coding on the same chains showed 83--89\% per-sentence agreement, indicating that batching reduced API cost roughly tenfold without meaningful loss of reliability. The canonical code for each sentence was the majority vote across the three judges; sentences for which all three judges produced a different label (a 1-1-1 split) were excluded from the canonical analysis (2.5\% of sentences). The per-chain unit of analysis was the proportion of canonical sentences in each of the five categories. Figure~\ref{fig:chain-ai} shows one chain coded under this scheme as a qualitative anchor; Figure~\ref{fig:chain-sketch} demonstrates how this pattern extends to hand-drawn sketches, where quality-assessment content is reduced.

\paragraph{Perceptual accuracy and human annotation.} To assess whether the models correctly identified what each drawing depicted, we used a separate LLM judge (MiniMax M2.7 via OpenRouter, temperature = 0) to extract the primary object-level concepts from each reasoning chain. These concepts were compared against consensus labels collected as part of the original AuDrA archival dataset \citep{patterson2024audra} but not previously analyzed. Labels were generated through a multi-stage human procedure: MTurk workers independently labeled each drawing, after which a research assistant consolidated the responses into the three most frequent label categories per starting image.

Each chain was classified as a match (the model identified the same concept as the human label, allowing for synonyms, near-synonyms, and reasonable hyponyms or hypernyms), partial match, no match, model abstained, or human uninterpretable. The last category was reserved for drawings whose final consensus label was left blank because raters could not agree on what the drawing depicted. We refer to these as ``No-Consensus Drawings'' (NCDs; N = 82 unique drawings, 246 chains across three models). This analysis was restricted to the hand-drawn-sketch dataset, because the polished DALL-E images afford little ambiguity in object recognition.

\paragraph{LLM hallucination audit.} For the 246 NCD chains, we conducted a follow-up audit in which the same LLM judge re-evaluated each chain and classified the model's confidence in its proposed identification into one of three categories. Confident identification meant that the model asserted a specific object without qualification. Hedged identification meant that the model named a possible object while expressing uncertainty. Abstained meant that the model explicitly stated it could not identify the drawing or that the drawing was abstract or unclear.

\paragraph{Analytic approach.} For the reasoning-on vs. reasoning-off comparison, we computed Pearson correlations with the human criterion separately for each condition on shared items, and we computed within-model rating consistency between the two conditions to confirm that the same model assigns similar ratings to the same image regardless of whether it reasons explicitly. For the chain-validation analysis, we report Pearson correlations of the predicted rating with both the original LLM rating and the human criterion. For the process-coding analysis, we report inter-judge reliability at the sentence level (pairwise percent agreement, three-way unanimous agreement, and pairwise Cohen's kappa, both overall and per-category as one-vs-rest), per-chain category profiles (mean and standard deviation of the five proportions, both pooled and broken out by reasoning model), and predictive validity tests in which per-chain category proportions are correlated with absolute model--human disagreement |AI - H| and with signed (AI - H). The same predictive correlations were recomputed as partial correlations controlling for image-level edge density, to test whether chain-content effects are independent of the visual-complexity confound documented in Study 1. For the perceptual-accuracy analysis, we report the proportion of chains in each match category and the Pearson correlation between AI rating and human rating within each category. For the hallucination audit, we report the proportion of chains in each confidence category and the corresponding correlations with human ratings.

\subsection*{Results}
\paragraph{Reasoning does not improve alignment.} We first asked whether engaging the reasoning chain improved alignment with human ratings. Compared to no-reasoning, we found that the reasoning chain did not meaningfully improve alignment with human ratings (Table~\ref{tab:reasoning-on-off}). Across all six within-model, within-image comparisons (three reasoning models x two datasets), the maximum advantage for reasoning-on was a negligible +.012 (Kimi K2.5 on human drawings), while interestingly, the minimum cost was a substantial -.075 (Qwen 3.6 Plus on AI-generated images). For two of the three models (Kimi K2.5 and GLM-5v Turbo) reasoning was essentially neutral on both datasets. Qwen 3.6 Plus showed a consistent \textasciitilde{}.07 reduction with reasoning enabled on both datasets. This pattern is consistent with broader findings in the LLM-as-judge literature suggesting that chain-of-thought sometimes degrades performance on subjective evaluation tasks \citep[e.g.,][]{zheng2023judge}. Thus, the value of reasoning here is not better alignment with human ratings but rather in a transparent view into the evaluative process.

\begin{table}[!htbp]
\centering
\small
\caption{Reasoning-on vs. reasoning-off alignment with human ratings, within-model, within-image.}\label{tab:reasoning-on-off}
\begin{tabular}{@{}llccc@{}}
\toprule
Model & Dataset & r ON & r OFF & Delta \\
\midrule
Kimi K2.5 & AI-generated images & .648 & .636 & +.012 \\
GLM-5v Turbo & AI-generated images & .570 & .572 & -.002 \\
Qwen 3.6 Plus & AI-generated images & .562 & .637 & -.075 \\
Kimi K2.5 & Hand-drawn sketches & .544 & .550 & -.005 \\
GLM-5v Turbo & Hand-drawn sketches & .444 & .462 & -.018 \\
Qwen 3.6 Plus & Hand-drawn sketches & .470 & .535 & -.065 \\
\bottomrule
\end{tabular}
\end{table}

\paragraph{Within-model rating consistency.} Before turning to the chain content, we verified that the same model assigns similar ratings to the same image regardless of whether it reasons explicitly. Reasoning-on and reasoning-off ratings were strongly positively correlated within each model overall. However, ratings were not perfectly correlated, and were notably reduced, particularly in the line-drawings dataset: Kimi K2.5 (AI images r = .76, human sketches r = .67), Qwen 3.6 Plus (AI images r = .75, human sketches r = .55), and GLM-5v Turbo (AI images r = .68, human sketches r = .51), confirming that the alignment differences reported above reflect some reweighting of evaluative criteria rather than fundamentally different ratings.

\paragraph{Chain language carries evaluative signal without the image.} To test whether the chains carry the evaluative signal that drives the ratings, we asked a separate text-only model (GPT-5.4 Nano) to predict each rating from the chain alone, with no access to the image and with all numeric digits removed from the chain text. We found the model recovered the original LLM rating with high accuracy: r = .82 between predicted and actual ratings overall (r = .81 for AI-generated images, r = .65 for hand-drawn sketches), with 69.8\% exact-match agreement and 99.4\% of predictions within +/-1 of the actual rating. Notably, the predicted ratings also correlated substantially with human ratings (r = .53 for AI-generated images, r = .40 for hand-drawn sketches), suggesting that the reasoning chains contain evaluative language aligned with human creative judgment.

\paragraph{Four evaluative dimensions structure chain content.} We next decomposed each chain into sentences and classified their evaluative content. The four evaluative dimensions were: Perception (describing visible content), Originality (judging novelty), Quality (judging visual aesthetic and composition), and Justification (reasoning about the 1-to-5 scale and committing to a rating). Three independent LLM judges classified each of 562,053 sentences across 7,475 chains into one of four evaluative categories or an Other residual bin. Coverage was high and stable across both datasets: 97.3\% of AI-generated images sentences and 97.6\% of human-drawn sketches sentences received a majority-vote canonical code, with the remainder reflecting 1-1-1 splits across the three judges (2.7\% and 2.4\% respectively). Inter-judge reliability was substantial across both datasets, with pairwise Cohen's kappa ranging from +.69 to +.79 on AI-generated images and from +.71 to +.81 on human-drawn sketches. Per-category kappa ranged from +.64 to +.76 on AI images and from +.53 to +.81 on human sketches, with the lowest values on Quality, which was also the rarest category on the sketches (only 3\% of sentences).

\begin{figure}[!ht]
\centering
\includegraphics[width=\linewidth]{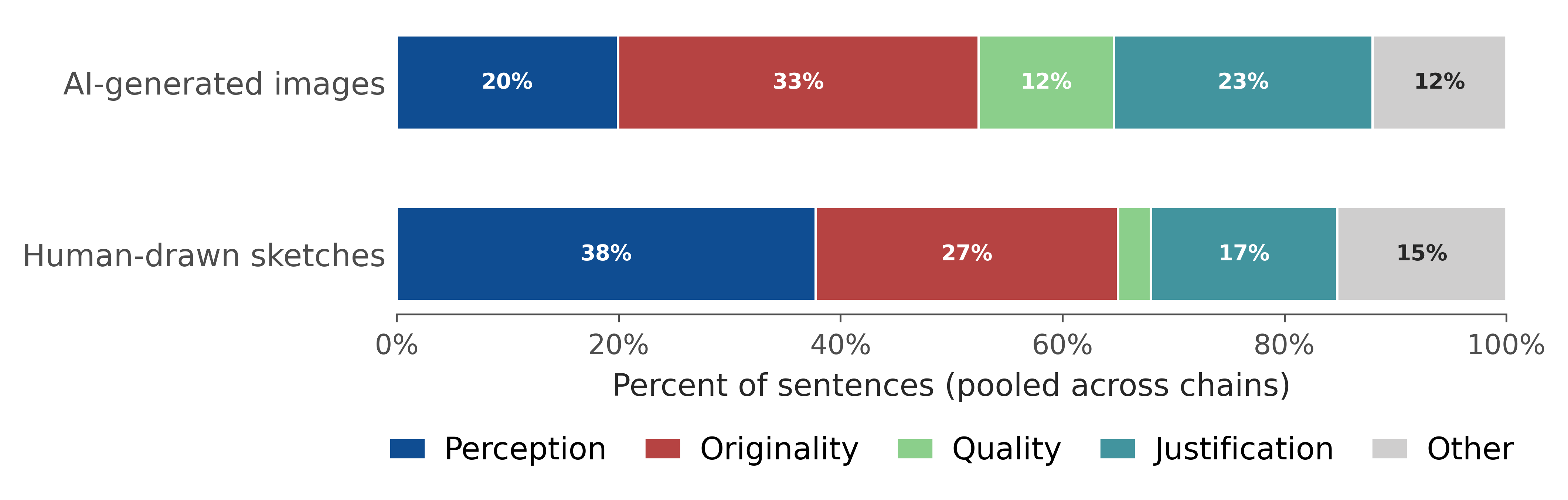}
\caption{Reasoning-chain content by dataset. Each sentence was classified into one of four evaluative categories (Perception, Originality, Quality, Justification) or a residual Other category; bars show the pooled percentage of sentences in each category, summed across all chains and all three reasoning-capable models. Hand-drawn sketches elicited nearly twice as much Perception (38\% vs. 20\%) and a quarter as much Quality (3\% vs. 12\%) as AI-generated images.}\label{fig:process-distribution}
\end{figure}

The pooled sentence-level distribution differed sharply between datasets (Figure~\ref{fig:process-distribution}). On AI-generated images, models devoted 32.5\% of reasoning sentences to Originality, 23.3\% to Justification, 19.9\% to Perception, 12.2\% to Quality, and 12.1\% to Other. On hand-drawn sketches, the same models devoted 37.8\% to Perception (nearly double compared to AI images), 27.2\% to Originality, 16.8\% to Justification, 15.3\% to Other, and only 3.0\% to Quality. These results suggest that the four categories capture stable evaluative tendencies of each LLM, and that models spend more or less time in different stages depending on the visual stimulus.

\begin{figure}[!ht]
\centering
\includegraphics[width=\linewidth]{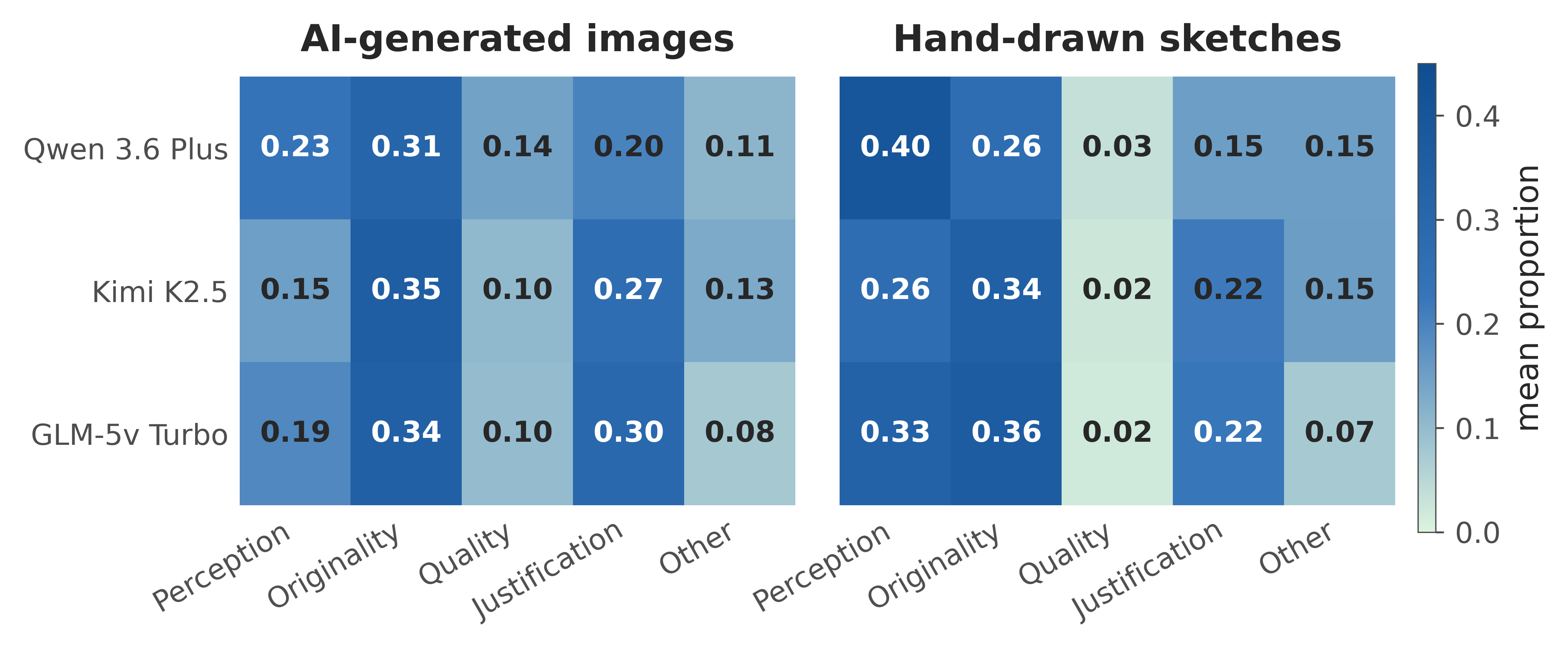}
\caption{Reasoning-chain content by model. Cells show the mean proportion of each chain in each category, averaged within model and dataset. Profiles are consistent across datasets: Qwen 3.6 Plus is most perception- and quality-heavy, GLM-5v Turbo most justification-heavy, and Kimi K2.5 produces the most Originality. Evaluative tendencies are thus model-specific rather than stimulus-driven.}\label{fig:process-heatmap}
\end{figure}

Regarding model effects, we found the three LLMs showed distinct evaluative profiles that held across both datasets (Figure~\ref{fig:process-heatmap}). Qwen 3.6 Plus was the most perception-heavy and quality-heavy on both datasets; GLM-5v Turbo was the most justification-heavy on both; and Kimi K2.5 produced the most Originality and `Other' content on human sketches, reflecting more transitions, hedging, and meta-commentary in its reasoning.

\begin{figure}[!ht]
\centering
\includegraphics[width=\linewidth]{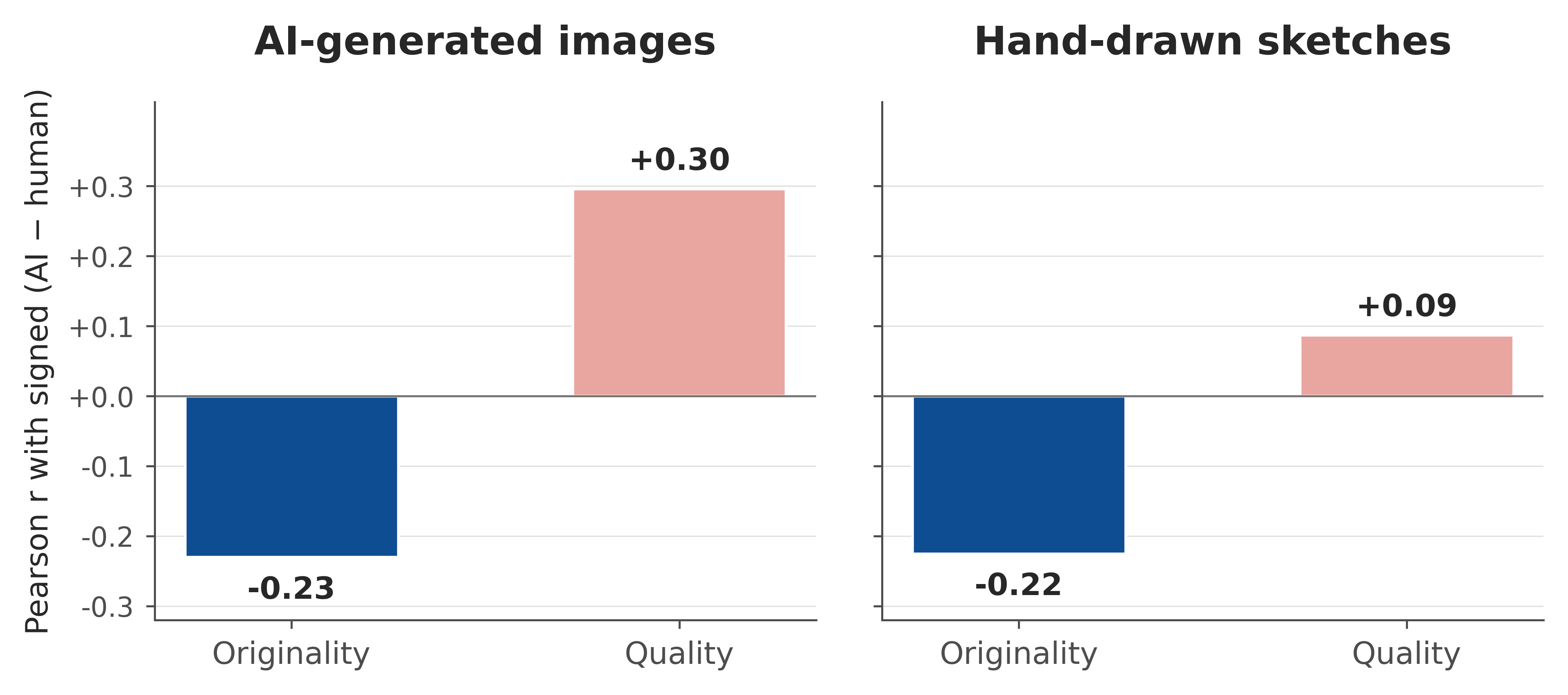}
\caption{How reasoning-chain content relates to model–human rating gaps. Bars show the Pearson correlation between a chain's proportion of Originality (or Quality) sentences and the signed model-minus-human rating difference: positive (coral) means the category pushes model ratings above humans', negative (blue) below. In both datasets, more Originality  is associated with harsher model ratings and more Quality  with more lenient ratings; the Quality effect is strongest on AI-generated images, where models were more lenient relative to human raters.}\label{fig:process-predictive}
\end{figure}

\paragraph{Originality and quality predict model--human divergence.} We next asked whether the four evaluative dimensions predicted where models diverged from human raters (Figure~\ref{fig:process-predictive}). Across both datasets, more Originality content was associated with lower model ratings relative to humans, while more Quality content was associated with higher model ratings. However, the Quality effect was substantially stronger on AI-generated images, where models were systematically lenient, than on human-drawn sketches, where Quality content was rare. Perception and Justification showed smaller, dataset-dependent effects, and Other was negligible. The same pattern held when we substituted absolute disagreement for signed disagreement. These results suggest that the categories driving model--human divergence are the same ones models spend the most evaluative effort on, with Originality consistently pulling ratings down and Quality pulling them up when models have polished images to evaluate.

\paragraph{Models identify objects but misidentification barely hurts alignment.} The reasoning chains describe not only evaluative criteria but also what each model believes the drawing depicts (i.e., Perception), which we compared against the human consensus labels using a separate LLM judge (Table~\ref{tab:perceptual-accuracy}). We found that excluding the drawings that humans themselves could not identify (NCD; see below), the models correctly identified the depicted object in 65.1\% of chains (2,770/4,253 match or partial match). Identification accuracy varied systematically with item difficulty: shape 4 yielded the highest match rate (75.6\%), followed by shape 11 (63.9\%), while shape 12 acted as a difficulty ceiling at 55.5\%. This gradient mirrors the per-shape pattern observed in Study 1.

The central question, however, was whether object misidentification undermines alignment with human creativity ratings. Results suggest that it did not: the correlation between model and human ratings for correctly identified drawings (r = .49, N = 2,605) was only marginally higher than for misidentified drawings (r = .43, N = 1,422). Partial matches (r = .43, N = 165) and explicit abstentions (r = .45, N = 61) fell within the same narrow band, suggesting that the creativity signal these models detect operates independently of whether the model correctly perceives what the drawing depicts.

\begin{table}[!htbp]
\centering
\small
\caption{Perceptual accuracy and rating alignment by match category (hand-drawn sketches).}\label{tab:perceptual-accuracy}
\begin{tabular}{@{}lll@{}}
\toprule
Match category & N & r with human rating \\
\midrule
Match & 2,605 & .49 \\
Partial match & 165 & .43 \\
No match & 1,422 & .43 \\
Model abstained & 61 & .45 \\
NCD (human uninterpretable) & 246 & .55 \\
\bottomrule
\end{tabular}
\end{table}

\paragraph{Alignment is strongest on drawings humans cannot identify.} Finally, we examined how models handle drawings that humans could not categorize. Eighty-two of the 1,500 hand-drawn sketches lacked a consensus label because raters could not agree on what the drawing depicted. These NCDs produced the highest model--human alignment of any match category (r = .55, N = 246 chains across three models). To characterize how the models handled these ambiguous stimuli, we classified each NCD chain by the model's confidence in its proposed identification. Most chains reflected hedged identification (63.4\%), in which the model named a possible object while expressing uncertainty; a smaller proportion reflected confident identification (33.3\%), in which the model asserted a specific object without qualification; and very few reflected outright abstention (3.3\%). Alignment with human ratings was robust across confidence categories (hedged: r = .56; confident: r = .51). These results suggest that when a drawing resists categorization, i.e., when neither the human nor the model can settle on a conventional label, both fall back on structural and compositional properties of the drawing itself, and these evaluations tend to converge.

\section*{Discussion}
The present research provides the first evidence that general-purpose multimodal LLMs can match human judgments of visual creativity without any additional training. Study 1 established that six models show strong positive correlations with human creativity ratings on both AI-generated images and hand-drawn sketches, with partial correlations confirming that alignment is not explained by low-level visual complexity. Study 2 demonstrated that the ``chains of thought'' produced by three reasoning models follow a four-part structure (perception, originality, quality, justification) that recurs across models and explains their final rating, providing an interpretable way to understand how AI models evaluate creativity.

In Study 1, Gemini 3 Flash showed the strongest alignment with human ratings across both datasets (r = .68 on each). Notably, two open-source models (Kimi K2.5, Qwen 3.6 Plus) outperformed a leading proprietary LLM (GPT-5.4 Mini), which produced the weakest alignment overall, particularly on human-drawn sketches (r = .29). Alignment was also weaker overall on sketches than on AI-generated images, even for the strongest models, with the rank-ordering of models shifting between the sketches and images. These results suggest that zero-shot scoring of visual creativity is feasible across a range of multimodal LLMs, but that model selection matters more for sparse stimuli (e.g., human sketches) than for polished ones (AI-generated images).

Our results extend the automated creativity scoring literature, which has focused heavily on verbal tasks, such as the AUT \citep{beaty2021semdis,organisciak2023beyond,saretzki2025german}. Such work typically fine-tunes language models on thousands of creative responses and human ratings, yielding a trained model that performs well on a specific creativity task (but not others). Recently, researchers have begun to explore the potential of LLMs for automated scoring using prompting alone, showing remarkably high correlations with human ratings without any examples \citep{organisciak2023beyond,saretzki2025german}. The current study extends this work to the figural domain, which has so far only fine-tuned vision classifiers requiring large-scale human-rated data \citep{acar2025figural,cropley2025tctdp,cropley2022cnn,patterson2024audra}. We achieve comparable correlations without task-specific training, using zero-shot prompting with only a brief scoring prompt given to human raters. Although our results point to the potential of zero-shot creativity scoring, we caution against using LLMs for this purpose on new tasks that have not yet been validated.

We observed two clear biases in LLM creativity ratings: an elaboration bias and a leniency bias. Regarding elaboration, models (and humans) tend to give higher ratings to more complex responses on both verbal and visual tasks \citep{domanti2026elaboration,patterson2024audra}: features like ink on the page are trivial for AI models to detect (i.e., more lines, higher rating). Crucially, in our study, LLM alignment with humans remained robust after controlling for edge density, suggesting they track some higher-level aspects of creativity. Regarding leniency bias, models gave higher creativity ratings on average to AI-generated images compared to human-drawn sketches, suggesting that LLM judges penalize the simplicity of line drawings despite prompt instructions to focus on originality, not technical proficiency. This pattern aligns with growing evidence of self-preference and familiarity biases in LLM-as-a-judge systems, where models assign disproportionately high ratings to outputs that resemble their own generations or fall within their training distribution \citep{panickssery2024self,wataoka2024selfpref}. This asymmetry is a considerable threat to validity of automated scoring that should be addressed in future work through careful calibration.

To our knowledge, this study is the first analysis of how LLMs reason about visual creativity, analyzing their ``thought process'' leading up to their final rating. We identified four stages of the evaluative process---image perception, originality assessment, quality assessment, and rating justification---that recurred across three different LLMs for both drawings and images. These reasoning stages are partly expected, however, given that reasoning models are post-trained to follow the broader perception $\rightarrow$ reasoning $\rightarrow$ integration structure identified in recent vision-language chain-of-thought work \citep{avogaro2026sparc,jiang2025vlmr3}. The value of analyzing reasoning traces is that they show how this general reasoning structure is instantiated in creative evaluation: models first describe the visual content, then evaluate originality and quality, and finally integrate these considerations into a rating.

Notably, reasoning did not improve alignment with human ratings relative to a no-reasoning baseline, consistent with prior work showing that reasoning does not necessarily improve performance on subjective evaluation tasks \citep{zheng2023judge}. Reasoning traces were therefore most useful as a source of interpretability, revealing systematic differences in how LLMs evaluated different visual inputs. Specifically, models spent more time on perception when evaluating hand-drawn sketches, but more time on quality when evaluating AI-generated images. Analyzing reasoning chains also clarified where model judgments diverged from human ratings, with originality language associated with lower ratings and quality language associated with higher ratings, particularly for AI-generated images. Together, these patterns suggest that reasoning traces can help audit LLM creativity judgments by showing not only whether models agree with humans, but which evaluative criteria appear to drive that agreement.

The perceptual-accuracy analysis further suggests that LLM creativity judgments are not reducible to object recognition. Although models correctly identified the depicted objects within human-drawn sketches in most cases, misidentification only modestly reduced alignment with human ratings. A particularly striking result came from drawings that human raters themselves could not clearly categorize, which showed the strongest model--human alignment of any match category. This pattern suggests that, when object identity is ambiguous, both humans and models may rely more heavily on broader visual features, such as elaboration, composition, and the integration of the starting shape. Thus, the creativity signal detected by LLMs appears partly independent of successful object recognition.

Several limitations are worth noting, particularly around the generalizability of the present findings. The two datasets differed in stimulus type, rating instructions, and human-scoring procedures, which makes direct comparisons between AI-generated images and hand-drawn sketches challenging. This was partly intentional, however, because the goal was to test whether zero-shot scoring generalizes across distinct forms of visual creativity. Future work should use more standardized designs to isolate stimulus effects from scoring-method effects. Our process analyses were also limited by the available data. Perceptual accuracy could only be tested for the hand-drawn sketches, the subset of drawings without consensus labels was relatively small, and the hallucination audit relied on a single LLM judge. Finally, the study focused on two visual creativity tasks. Broader claims about visual creativity assessment will require testing additional domains, such as design, photography, and mixed-media work, ideally with calibrated scoring procedures and human benchmarks collected under common instructions.

In sum, the present findings show for the first time that multimodal LLMs can match human judgments of visual creativity. We find their reasoning follows a consistent and interpretable sequence that helps explain how models arrive at their ratings. Moreover, model--human alignment remained robust to object misidentification and persisted after controlling for visual complexity. Multimodal LLMs thus offer a promising tool for visual creativity assessment, while also making their biases easier to detect, quantify, and mitigate. To support secondary use of this pipeline by other researchers, we also share an open web app that runs the same scoring procedure on user-supplied images (Appendix~A).

\section*{Author Note}

Correspondence concerning this article should be addressed to Roger E. Beaty, Department of Psychology, The Pennsylvania State University, University Park, PA 16802. Email: rebeaty@psu.edu.

\paragraph{Use of Generative AI.} The authors used Anthropic's \texttt{claude-opus-4-7} and Claude Code to assist with data processing, generating analysis code, and manuscript preparation. All AI-assisted content, code, and outputs were reviewed, verified, and edited by the authors, who take full responsibility for the accuracy of the final manuscript.

\paragraph{Data Availability.} All data and analysis code are available at \url{https://anonymous.4open.science/r/visual-creativity-llm-judge-anonymous-8D7D}. An interactive scoring app that reproduces the pipeline on user-supplied images is hosted at \url{https://review-visual-eval-scoring.hf.space/} (a brief tutorial appears in Appendix~A). Raw stimuli are available through the original datasets \citep{orwig2026aiart,patterson2024audra}.

\paragraph{Author Contributions.} R.E.B.: conceptualization, methodology, software, formal analysis, visualization, writing -- review \& editing, supervision. W.O.: investigation (collected the AI-generated image dataset), writing -- original draft. Both authors approved the final version.

\paragraph{Acknowledgments.} R.E.B. was supported by Amazon and by the National Science Foundation under Grants DRL-2400782 and DUE-2155070.

\paragraph{Open Practices.} The present research involved secondary analysis of two previously published datasets \citep{orwig2026aiart,patterson2024audra}; no new data were collected. All model ratings, reasoning chains, process-coding outputs, and analysis code are openly available at the GitHub repository above. The study was not preregistered.

\paragraph{Competing Interests.} The authors declare no competing interests.

\bibliographystyle{apacite}
\bibliography{references}

\clearpage
\appendix
\section*{Appendix A: Using the Scoring App}
\addcontentsline{toc}{section}{Appendix A: Using the Scoring App}

To make the pipeline accessible beyond the analyses reported here, we built a small web app that reproduces the manuscript's scoring procedure on any user-supplied image. The app runs in a browser, requires no installation, and is hosted at \url{https://review-visual-eval-scoring.hf.space/}. Source code lives in the \texttt{scoring\_app/} directory of the GitHub repository.

\paragraph{Setup.} The app routes every model call through OpenRouter,\footnote{\url{https://openrouter.ai/}} a unified OpenAI-compatible API across model providers, so users need only a single API key. A key can be obtained at \url{https://openrouter.ai/keys}. The key is sent only to OpenRouter for the requests the user triggers and is not logged, stored, or persisted by the app. Source code is publicly available for inspection.

\begin{figure}[!ht]
\centering
\includegraphics[width=\linewidth]{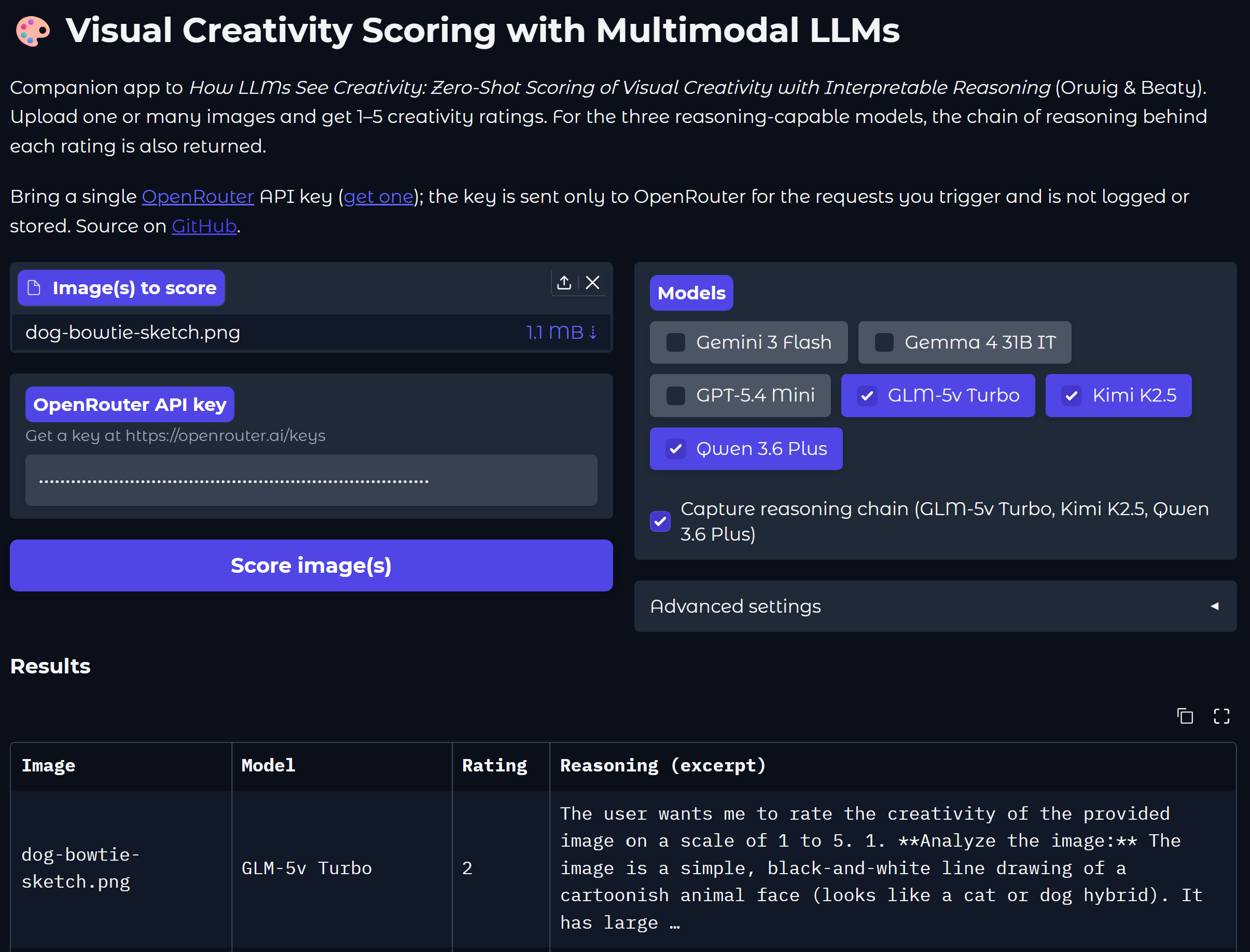}
\caption{Screenshot of the scoring app. The user uploads one or more images, supplies a single OpenRouter API key, and recovers per-model creativity ratings (and, for reasoning-capable models, the underlying reasoning chains) using the exact prompts from the manuscript.}\label{fig:app-screenshot}
\end{figure}

\paragraph{Scoring a single image.} Users drag-and-drop one image into the upload box, paste their OpenRouter key, and click \emph{Score image(s)}. By default, all six models are queried and the reasoning chain is captured for the three reasoning-capable models (GLM-5v Turbo, Kimi K2.5, Qwen 3.6 Plus). Ratings populate a four-column table (image filename, model, 1--5 rating, and a short excerpt of the reasoning chain). The non-reasoning models typically return a rating within a few seconds, while the reasoning-capable models take noticeably longer, since they generate the full reasoning chain before committing to a final rating. A representative session is shown in Figure~\ref{fig:app-screenshot}.

\paragraph{Scoring a batch.} The same upload box accepts any number of images at once. Calls run in parallel and retry automatically on transient errors, so a batch finishes substantially faster than scoring images one at a time.

\paragraph{Customizing the scoring prompt.} The app ships with the AI-generated-image prompt loaded by default. \emph{Advanced settings} exposes the full prompt text in an editable box, along with one-click buttons to load either the AI-image prompt or the hand-drawn-sketch prompt from the manuscript verbatim. Researchers who wish to apply this pipeline to a new visual domain can edit the prompt directly without changing any code; the new prompt is then used for every model on every image in the batch.

\paragraph{Other adjustable settings.} \emph{Temperature} (default 0; matches the manuscript), maximum output tokens, and the level of concurrency for parallel API calls are exposed as sliders in \emph{Advanced settings}.

\paragraph{Output.} After each run, the results table shows one row per (image, model) pair. A downloadable CSV preserves the full reasoning text for every cell, with columns \texttt{image}, \texttt{model}, \texttt{rating}, and \texttt{reasoning}. The CSV is regenerated on each scoring run; users should save it locally before scoring a new batch.

\paragraph{Reproducibility.} The app uses exactly the model identifiers listed in Table~\ref{tab:models} and the prompts in Section 2.3. Users who set temperature to zero will recover the same ratings reported in the manuscript when scoring the same images on the same model checkpoints.

\end{document}